\documentclass[pdflatex,sn-mathphys-num]{sn-jnl}

\usepackage{array}
\usepackage{graphicx}%
\usepackage{multirow}%
\usepackage{amsmath,amssymb,amsfonts}%
\usepackage{amsthm}%
\usepackage{mathrsfs}%
\usepackage[title]{appendix}%
\usepackage{xcolor}%
\usepackage{textcomp}%
\usepackage{manyfoot}%
\usepackage{colortbl}
\usepackage{makecell}
\usepackage{booktabs}%
\usepackage{algorithm}%
\usepackage{algorithmicx}%
\usepackage{algpseudocode}%
\usepackage{listings}%
\usepackage{adjustbox}
\usepackage{caption}
\theoremstyle{thmstyleone}%
\theoremstyle{thmstyletwo}%

\theoremstyle{thmstylethree}%

\begin{document}

\title[SurgAnt-ViVQA: Learning to Anticipate Surgical Events through GRU-Driven Temporal Cross-Attention]{SurgAnt-ViVQA: Learning to Anticipate Surgical Events through GRU-Driven Temporal Cross-Attention}

\author[1,2]{\fnm{Shreyas C.} \sur{Dhake}}
\author[1,2]{\fnm{Jiayuan} \sur{Huang}}
\author[1,2]{\fnm{Runlong} \sur{He}}
\author[3,4]{\fnm{Danyal Z.} \sur{Khan}}
\author[1,2]{\fnm{Evangelos B.} \sur{Mazomenos}}
\author[1,3]{\fnm{Sophia} \sur{Bano}}
\author[1,4]{\fnm{Hani J.} \sur{Marcus}}
\author[1,3]{\fnm{Danail} \sur{Stoyanov}}
\author[1,2]{\fnm{Matthew J.} \sur{Clarkson}}
\author[1,2,5]{\fnm{Mobarak I.} \sur{Hoque}}

\affil[1]{\orgdiv{UCL Hawkes Institute}, 
            \orgname{University College London}, 
            \orgaddress{\country{UK}}}
\affil[2]{\orgdiv{Dept of Medical Physics \& Biomedical Engineering}, 
           \orgname{UCL}, 
           \orgaddress{\country{UK}}}
\affil[3]{\orgdiv{Dept of Computer Science}, 
\orgname{University College London}, 
\orgaddress{\country{UK}}}
\affil[4]{\orgdiv{Dept of Neurosurgery}, 
           \orgname{National Hospital for Neurology and Neurosurgery}, 
           \orgaddress{\country{UK}}}
\affil[5]{\orgdiv{Division of Informatics, Imaging and Data Science}, \orgname{The University of Manchester}, \orgaddress{\country{UK}}}

\abstract{
\textbf{Purpose:} Anticipating forthcoming surgical events is vital for real-time assistance in endonasal transsphenoidal pituitary surgery, where visibility is limited and workflow changes rapidly. Most visual question answering (VQA) systems reason on isolated frames with static vision language alignment, providing little support for forecasting next steps or instrument needs. Existing surgical VQA datasets likewise center on the current scene rather than the near future.

\textbf{Methods:} We introduce PitVQA-Anticipation, the first VQA dataset designed for forward looking surgical reasoning. It comprises 33.5 hours of operative video and 734,769 question answer pairs built from temporally grouped clips and expert annotations across four tasks: predicting the future phase, next step, upcoming instrument, and remaining duration. We further propose SurgAnt-ViVQA, a video language model that adapts a large language model using a GRU Gated Temporal Cross-Attention module. A bidirectional GRU encodes frame to frame dynamics, while an adaptive gate injects visual context into the language stream at the token level. Parameter efficient fine tuning customizes the language backbone to the surgical domain.

\textbf{Results:} SurgAnt-ViVQA tested upon on PitVQA-Anticipation (BLEU-4 72.38, ROUGE-L 84.94, METEOR 87.05) and EndoVis datasets, surpassing strong image and video based baselines. Ablations show that temporal recurrence and gated fusion drive most of the gains. A frame budget study indicates a trade-off: 8 frames maximize fluency, whereas 32 frames slightly reduce BLEU but improve numeric time estimation.

\textbf{Conclusion:} By pairing a temporally aware encoder with fine grained gated cross-attention, SurgAnt-ViVQA advances surgical VQA from retrospective description to proactive anticipation. PitVQA-Anticipation offers a comprehensive benchmark for this setting and highlights the importance of targeted temporal modeling for reliable, future aware surgical assistance.
}

\maketitle
\section{Introduction}

Anticipating future events in surgery is a critical capability for building real-time, systems achieving this can help scrub nurses in preparation (e.g. instrument anticipation) or scheduling of operating rooms (OR). Surgical procedures such as endonasal transsphenoidal pituitary surgery, where surgeons must operate within a narrow corridor and interpret complex, occluded anatomy with limited field of view~\cite{khan2023current}, anticipation is essential. While recent advancements in Visual Question Answering (VQA) offer promising tools for surgical guidance, most existing methods focus on static frame-level reasoning, lacking temporal modeling required to understand surgical progressions and predict upcoming events. Furthermore, existing surgical VQA datasets are primarily designed for retrospective queries grounded in the current scene, offering limited utility for anticipatory decision support.

Several recent works have advanced surgical VQA by building dedicated datasets~\cite{seenivasan2022surgical,he2025pitvqa++} and developing vision-language models~\cite{ he2024pitvqa} that address question–answer pairs related to tool presence, surgical phases, anatomical recognition, spatial relationships, surgical actions, workflow understanding, and tool activity. However, these models typically operate on isolated frames or short clips and are limited to answering questions based on the current or past visual context. None of them address the critical task of anticipatory reasoning in surgery, nor do they leverage full-length videos for comprehensive temporal question answering.

In parallel, significant progress has been made in modeling temporal information using recurrent networks such as LSTMs and GRUs~\cite{traore20202d}, as well as transformer-based models like Video ViT~\cite{arnab2021vivit}, Video Swin Transformer~\cite{liu2022video}, and XCLIP~\cite{ma2022x}. These models capture long-range dependencies across video frames and have been applied to tasks such as action recognition and video captioning. Recent vision-language models (VLMs), including Video-LLaMA3 ~\cite{zhang2025videollama}, have introduced cross-attention mechanisms to fuse temporal visual features with language tokens, yet they remain underexplored for fine-grained, time-sensitive video-text reasoning. While Video-LLaMA3 shows promise, it lacks fine control over temporal granularity and performs suboptimally on domain-specific anticipatory tasks such as surgery.

To bridge this gap, current research was done to anticipate phases, instruments or duration but none explore VQAs \cite{yuan2022anticipation,boels2025swag}. We introduce PitVQA-Anticipation, the first video-based surgical VQA dataset explicitly designed for forward-looking reasoning in pituitary surgery. Covering 33.5 hours of high-resolution surgical videos and comprising 734,769 question–answer (QA) pairs, the dataset spans four anticipatory tasks: forecasting future surgical phases, next operative steps, upcoming instrument usage, and remaining duration. Each QA pair is constructed from temporally grouped frames and enriched with expert surgical annotations to enable robust training of temporally-aware models.

In addition to the dataset, we propose SurgAnt-ViVQA, a novel video-language architecture for anticipatory surgical understanding. At its core is a GRU-Gated Temporal Cross-Attention module that models fine-grained temporal dynamics and adaptively integrates visual information into the language stream, enabling context-aware multimodal reasoning. Our main contributions are:

\begin{itemize}
\item \textbf{PitVQA-Anticipation Dataset:} A large-scale, expert-annotated video-based VQA dataset designed specifically for anticipatory surgical tasks, covering future phase, step, instrument, and duration prediction.

\item  \textbf{SurgAnt-ViVQA Model:} A novel video-language architecture featuring a GRU-Gated Temporal Cross-Attention mechanism for dynamic and fine-grained integration of visual context into language representations.

\item \textbf{Benchmarking and Evaluation:} Comprehensive experiments on two datasets showing that SurgAnt-ViVQA outperforms strong baselines and existing state-of-the-art models on all anticipatory tasks, setting a new benchmark for future-aware surgical AI.
\end{itemize}

This work takes a significant step toward enabling predictive reasoning in surgery, with implications for proactive decision support and real-time surgical assistance.

\begin{figure}[!h]
    \centering
    \includegraphics[width=1\linewidth]{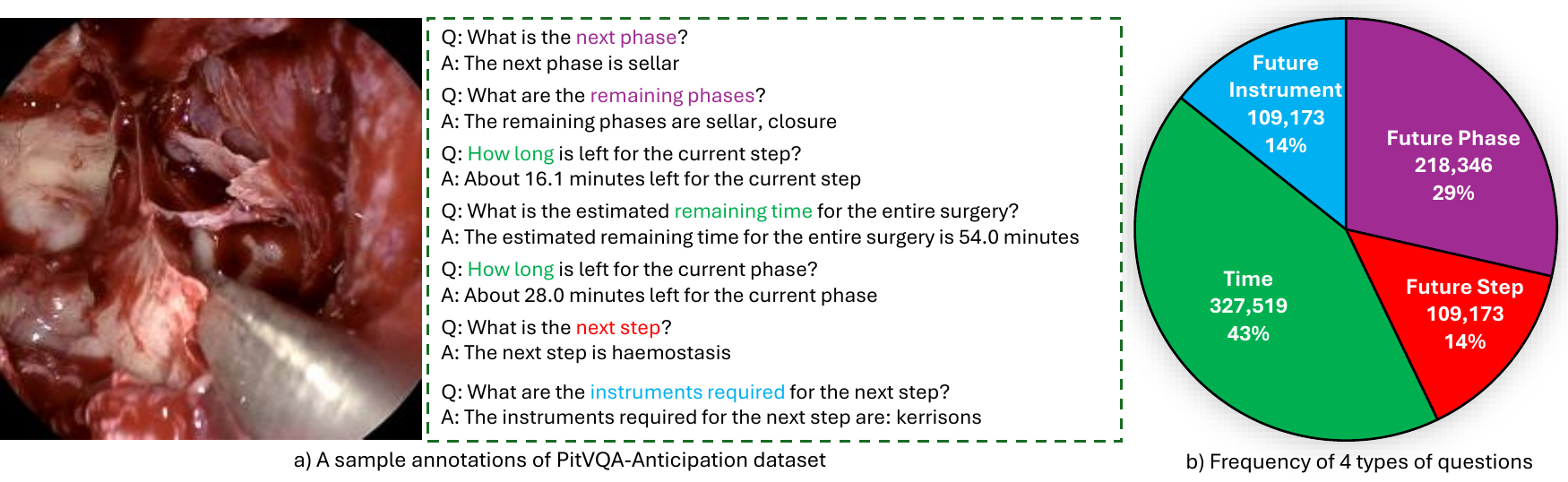}
    \caption{Sample question–answer (QA) pairs from the PitVQA-Anticipation dataset displays anticipatory queries regarding upcoming surgical phases, steps, instruments, and time durations.}
    \label{fig:dataset_stats}
\end{figure}

\section{Method}
\subsection{Proposed Dataset: PitVQA-Anticipation}

Our PitVQA-Anticipation dataset uses PitVis dataset \cite{das2024pitvis,he2024pitvqa} which is made up of 25 endoscopic pituitary surgery videos from National Hospital of Neurology and Neurosurgery in London, with a total duration of approximately 33.5 hours. The surgeries were recorded using high-definition endoscopes (Karl Storz Endoscopy) at 720p resolution and stored as MP4 files. All videos were annotated by two neurosurgical residents with pituitary surgery experience following a standardized annotation framework for surgical phases, steps, present instruments, and surgical activities, with supervision from an attending neurosurgeon. We extracted frames at 1 FPS and removed any blurred or occluded frames based on the annotations. Ultimately, we obtained a total of 109,173 frames.

We generated QA pairs for each frame using templates reviewed by surgical residents for 4 categories of questions, including future surgical phases, steps, instruments, and remaining time left. Fig~\ref{fig:dataset_stats}.(a) shows an example of QA pairs from our PitVQA-Anticipation dataset. These questions are based on ongoing pituitary surgeries and address concerns of surgeons, anaesthetists, and nurses, such as remaining time and instrument preparation. Fig~\ref{fig:dataset_stats}.(b) illustrates the distribution of the 4 question categories, with time-related questions comprising the largest proportion at approximately 43\%. The annotations encompass 40 categories in total, including 15 future steps, 4 future phases, 18 future instruments, and 3 types of time-related annotations, as detailed in Table~\ref{table:dataset_table}.

To enhance temporal understanding and provide richer contextual information for surgical anticipation, we grouped every 8 consecutive frames into a single clip, using the QA pairs from the last frame of each clip as the corresponding QA pairs for that sample. To ensure consistent temporal context, we discarded any clip that contained missing or invalid frames, resulting in a total of 734,769 QA pairs derived from 8-frame clips. For evaluation, we randomly selected 5 videos ([‘02’, ‘06’, ‘12’, ‘13’, ‘24’]) as the test set, comprising 166,229 QA pairs (22.63\%), while the remaining 20 videos were used for training, containing 568,540 QA pairs (77.37\%). 

\begin{figure*}[t]
    \centering
    \includegraphics[width=\linewidth]{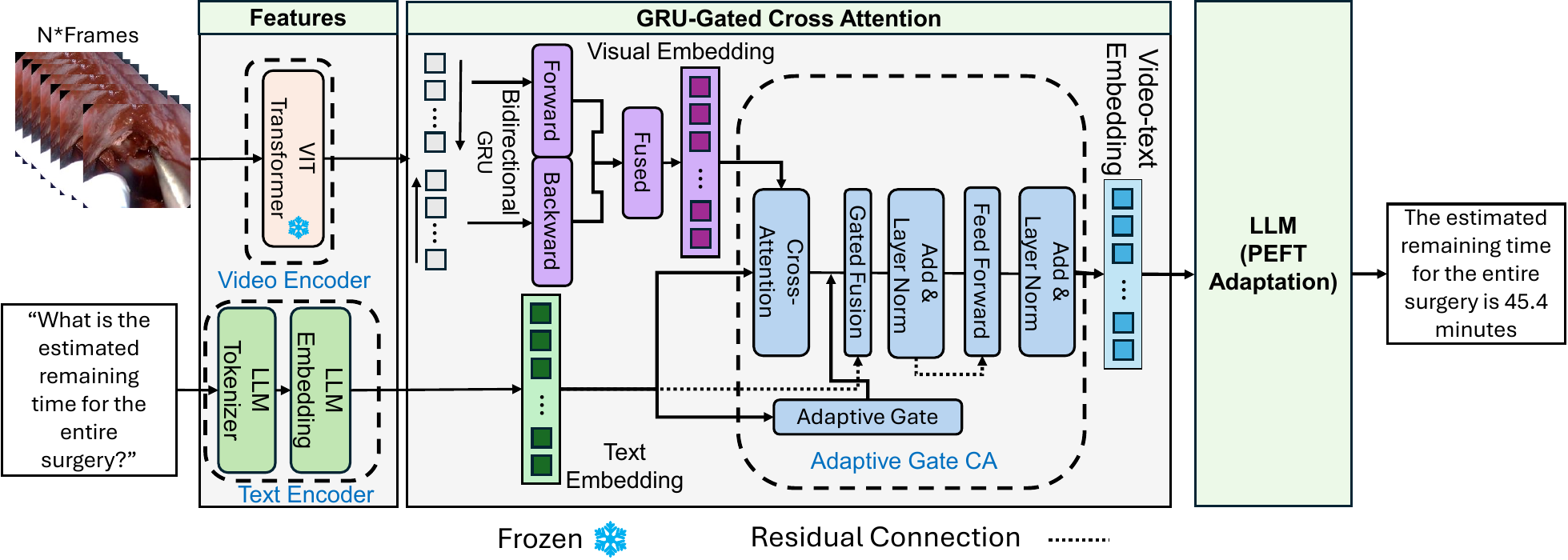}
    \caption{The SurgAnt-ViVQA architecture processes text through the LLM’s tokenizer and embeddings, while the GRU-Gated Cross-Attention module fuses video features before passing the combined representation to the LLM, adapted via parameter-efficient fine-tuning (LoRA) \cite{hu2022lora}.}
    \label{fig:model}
\end{figure*}

\begin{table}[h]
\centering
\caption{Categorical key annotations in our PitVQA-Anticipation dataset.}
\label{table:dataset_table}
\begin{tabular}{>{\centering\arraybackslash}m{2.2cm}|>{\centering\arraybackslash}m{0.6cm}|p{8cm}} 
\hline
\textbf{Category} & \textbf{No.} & \multicolumn{1}{c}{\textbf{Class Annotation}} \\ \hline
\begin{tabular}[c]{@{}c@{}}Future\\Steps\end{tabular} & 15 &
Nasal corridor creation, anterior sphenoidotomy, septum displacement, sphenoid sinus clearance, sellotomy, haemostasis, synthetic graft placement, durotomy, tumour excision, fat graft placement, gasket seal construct, dural sealant, nasal packing, debris clearance, end of step \\ \hline
\begin{tabular}[c]{@{}c@{}}Future\\Phases\end{tabular} & 4 &
Nasal sphenoid, sellar, closure, end of phase \\ \hline
\begin{tabular}[c]{@{}c@{}}Future\\Instruments\end{tabular} & 18 &
Suction, freer elevator, pituitary rongeurs, spatula dissector, kerrisons, cottle, haemostatic foam, micro doppler, nasal cutting forceps, stealth pointer, irrigation syringe, retractable knife, dural scissors, ring curette, cup forceps, bipolar forceps, tissue glue, surgical drill \\ \hline
Time & 3 & In minutes for step, phase and overall surgery \\ \hline
\end{tabular}
\end{table}

\subsection{Proposed Method: SurgAnt-ViVQA}

We propose \textbf{SurgAnt-ViVQA} (Fig~\ref{fig:model}), a multimodal architecture for video-based visual question answering in pituitary surgery. Video frames are encoded through a ViT-based encoder followed by a bidirectional GRU that captures temporal dynamics. Text inputs are tokenized and embedded using a GPT-2 language model. A novel cross-modal module, the \textit{GRU-Gated Cross-Attention}, fuses video and text features via an adaptive gating mechanism. The fused representations are then decoded by a parameter-efficient GPT-2 fine-tuned with LoRA (Low-Rank Adaptation)~\cite{hu2022lora}.

\subsubsection{GRU-Gated Cross-Attention}
To integrate visual and textual modalities, we design a cross-attention block composed of three main components: (i) a \textbf{temporal encoder}, (ii) a \textbf{cross-modal attention} mechanism, and (iii) an \textbf{adaptive gating} layer.

\paragraph{Temporal Video Encoding.}
Given video frame features $X^v = [\mathbf{x}^v_1,\dots,\mathbf{x}^v_T] \in \mathbb{R}^{T \times D}$, a bidirectional GRU captures forward and backward dependencies:
\[
\mathbf{h}^v_t = \overrightarrow{\mathbf{h}}^v_t + \overleftarrow{\mathbf{h}}^v_t,
\]
producing context vectors $H^v = [\mathbf{h}^v_1,\dots,\mathbf{h}^v_T]$ that encode temporal information such as actions and transitions.

\paragraph{Cross-Modal Attention.}
Text embeddings $X^t = [\mathbf{x}^t_1,\dots,\mathbf{x}^t_L]$ query the video context $H^v$ through multi-head attention:
\[
A = \text{softmax}\!\Big( \frac{(X^t W^Q)(H^v W^K)^\top}{\sqrt{d_k}}\Big)\!(H^v W^V),
\]
where $W^Q$, $W^K$, and $W^V$ are projection matrices. Each $\mathbf{a}_i \in A$ represents the visual features most relevant to the $i$-th token.

\paragraph{Adaptive Gating and Fusion.}
To control the flow of visual information, we compute a gating vector for each text token:
\[
\mathbf{g}_i = \sigma(W_g \mathbf{x}^t_i + \mathbf{b}_g),
\quad
\tilde{\mathbf{a}}_i = \mathbf{g}_i \odot \mathbf{a}_i,
\]
where $\sigma$ is the sigmoid function. The gated outputs $\tilde{A}$ are added to $X^t$ and normalized:
\[
H^t = \text{LayerNorm}(X^t + \tilde{A}).
\]
Finally, a feed-forward network refines $H^t$ via another residual connection:
\[
Z = \text{LayerNorm}(H^t + \text{FFN}(H^t)),
\]
yielding fused embeddings $Z$ that integrate linguistic and visual cues for downstream decoding.

\subsubsection{Parameter-Efficient Fine-Tuning with LoRA}
To adapt the LLMs efficiently, we employ \textbf{Low-Rank Adaptation (LoRA)}~\cite{hu2022lora}. Instead of updating all model parameters, LoRA inserts trainable low rank matrices into specific projection layers (here, the attention and output projections: \texttt{c\_attn}, \texttt{c\_proj}). This greatly reduces the number of trainable parameters while maintaining full model performance. In SurgAnt-ViVQA, LoRA enables lightweight, domain specific fine-tuning of the GPT-2 language model to generate accurate surgical answers conditioned on fused visual text representations.

\section{Experimental Setup}
The tables that shall highlight the results will include the following short forms: Modality (Mod.), BLEU (B.-1 to B.-4), ROUGE-L (R.-L), and METEOR (MET.).
\begin{table*}[!t]
\centering
\caption{Performance comparison of SurgAnt-ViVQA with various baseline models using the PitVQA-Anticipation dataset. Our framework integrates two language backbones that both outperform image based and video language baselines, with SurgAnt-ViVQA (GPT-2) achieving the best overall performance.}
\label{tab:combination}
\scalebox{0.75}{
\begin{tabular}{c|c|c|c c c c c c}
\hline
\textbf{Strategy} & \textbf{Models} & \textbf{Mod.} &
\textbf{B.-1} & \textbf{B.-2} & \textbf{B.-3} & \textbf{B.-4} &
\textbf{R.-L} & \textbf{MET.} \\ \hline

\multirow{4}{*}{Zero Shot}
 & MedGemma~\cite{sellergren2025medgemma}       & Image & 27.09 & 20.17 & 15.03 &  9.16 & 41.37 & 47.16 \\
 & Qwen 2.5 VL~\cite{bai2025qwen2}              & Image & 33.88 & 26.89 & 21.43 & 16.35 & 47.11 & 52.88 \\
 & LLaMA-Vision 3.2~\cite{grattafiori2024llama} & Image & 16.48 & 11.18 &  8.36 &  6.01 & 37.67 & 45.49 \\
 & Video-LLaMA3~\cite{zhang2025videollama}      & Video & 41.41 & 31.40 & 25.67 & 19.76 & 51.42 & 55.50 \\ \hline

\multirow{2}{*}{Fully Finetuned}
 & VisualBert~\cite{li2019visualbert}           & Image & 18.27 & 16.46 & 16.46 & 14.58 & 12.61 & 62.84 \\
 & VB-RM~\cite{seenivasan2022surgical}          & Image & 19.56 & 17.64 & 15.61 & 13.50 & 33.02 & 58.82 \\ \hline

\multirow{3}{*}{PEFT}
 & PitVQA++~\cite{he2025pitvqa++}               & Image & 42.52 & 37.39 & 35.10 & 32.95 & 53.91 & 53.90 \\
 & SurgAnt-ViVQA (Qwen3-0.6B)                   & Video & 88.13 & 80.07 & 76.69 & 72.04 & 84.82 & 87.36 \\
 & \textbf{SurgAnt-ViVQA (GPT-2)}               & \textbf{Video} & \textbf{88.50} & \textbf{80.35} & \textbf{77.02} & \textbf{72.38} & \textbf{84.94} & \textbf{87.05} \\ \hline
\end{tabular}
}
\end{table*}

\bmhead{Dataset}In addition to PitVQA-Anticipation dataset we also evaluated our model upon public benchmark dataset of EndoVis18-VQA \cite{seenivasan2022surgical}. This dataset contains 13,790 question-answer pairs derived from 2,086 surgical scenes across 14 nephrectomy surgery videos with 50 unique words. We follow the original data split \cite{seenivasan2022surgical} for training and validation sets. The training set comprises 1,560 frames and 10,574 question-answer pairs, while the validation set consists of 447 frames and 3,216 question-answer pairs. 

\bmhead{Implementation Details}Experiments on the EndoVis18-VQA dataset were trained on an NVIDIA RTX A6000 GPU (48 GB) using 60 epochs, batch size = 32, and learning rate = 2e-5. Experiments on the PitVQA-Anticipation dataset were run on an NVIDIA RTX A100 GPU (80 GB) with the same learning rate and epochs but a batch size = 4, following the baselines in \cite{seenivasan2022surgical}. The EndoVis18-VQA split followed \cite{seenivasan2022surgical}, using subsets [1, 5, 16] for validation and the rest for training. For PitVQA-Anticipation, videos [02, 06, 12, 13, 24] were used for validation, with the remaining for training. We applied Low-Rank Adaptation (LoRA)~\cite{hu2022lora} to GPT-2’s attention and projection layers (\texttt{c\_attn}, \texttt{c\_proj}) with $r{=}8$, $\alpha{=}16$, and dropout{=}0.1, allowing efficient fine-tuning while keeping the base model frozen.
\section{Results and Discussion}

Table~\ref{tab:combination} presents a comparative study using our PitVQA-Anticipation dataset to evaluate how effectively various models can understand and anticipate future surgical events. Initially, we conducted zero shot evaluations on MedGemma, Qwen 2.5 VL, and LLaMA-Vision 3.2. These models demonstrated limited understanding when applied to surgery specific data, highlighting the necessity of domain specific fine-tuning for surgical VQA tasks. Among these, only Video-LLaMA3 is designed for video inputs, while the others operate on individual image frames. As expected, Video-LLaMA3 outperformed the image based models in the zero shot setting, likely due to its ability to leverage temporal information across frames. This underscores the importance of modeling sequential context in surgical videos, where understanding future events depends on the evolution of visual cues over time rather than static frame analysis. Fully fine-tuning VisualBert (VB), VRBM on the new dataset fares even worse, with BLEU-4 topping out at 14.58\%, whilst Meteor scores may seem high but is only due to the data being very noisy and repeating key words leading to deceivingly higher Meteor scores. Parameter efficient fine-tuning (PEFT) we used one a strong image VQA model, PitVQA++  which pushes BLEU-4 to 32.95\%. Nonetheless, our SurgAnt-ViVQA, is able to beat all the previous models by significant margins.

\begin{table*}[!t]
\centering
\caption{Comparison of five cross-attention (CA) variants on PitVQA-Anticipation.  
All baselines share an X-CLIP vision backbone; our variant replaces it with a GRU-gated temporal encoder.}
\label{tab:pit_res}
\scalebox{0.75}{
\begin{tabular}{l|c|c|c|c|c|c|c}
\hline
\textbf{Model} & \textbf{Temporal} & \textbf{B.-1} & \textbf{B.-2} & \textbf{B.-3} & \textbf{B.-4} & \textbf{R.-L} & \textbf{MET.} \\ \hline
Multistage CA\cite{shang2023vision}      & \multirow{4}{*}{X-CLIP\cite{ma2022x}} & 54.35 & 32.55 & 18.88 & 11.66 & 60.55 & 51.90 \\
FusionTok CA\cite{lee2024multi}                                       &  & 68.59 & 59.29 & 50.40 & 42.13 & 71.93 & 72.02 \\
LoRA-Enhanced CA\cite{gong2023multimodal}       &  & 34.69 & 20.71 & 13.10 &  9.21 & 37.96 & 35.48 \\
Dynamic-Gate CA\cite{praveen2024dynamic}     &  & 34.49 & 19.12 & 11.61 &  7.67 & 35.82 & 32.39 \\ 
\textbf{Gated Temporal CA (ours)}                                     & \textbf{GRU} & \textbf{88.50} & \textbf{80.35} & \textbf{77.02} & \textbf{72.38} & \textbf{84.94} & \textbf{87.05} \\ \hline
\end{tabular}
}
\end{table*}

Table~\ref{table:categorical_metrics} reports category wise results across the four anticipatory tasks in PitVQA-Anticipation. SurgAnt-ViVQA achieves the highest accuracy for future phase prediction (82.5 \% with GPT-2, 80.96 \% with Qwen3-0.6B), indicating strong temporal context modeling. Future step performs lowest which shows the model struggles to interpret what should follow whereas for instruments the model performs better despite the number of unique answers, while MAE values confirm accurate temporal estimation for both phase and step durations (13 min and 9 min). The comparable results between GPT-2 and the lighter Qwen3-0.6B backbone suggest that performance gains primarily arise from the GRU-Gated Temporal Cross-Attention rather than model scale.

\begin{table*}[!htb]
\centering
\captionsetup{aboveskip=3pt, belowskip=2pt}
\caption{Accuracy and MAE (Mean Absolute Error) metrics for each category upon PitVQA-Anticipation dataset.}
\label{table:categorical_metrics}
\scalebox{0.75}{
\begin{tabular}{c|ccc|ccc}
\hline
\multicolumn{1}{l|}{}                                                            & \multicolumn{3}{c|}{\textbf{Accuracy}}                                                                                                                                                                                                                                                                              & \multicolumn{3}{c}{\textbf{MAE}}                                                            \\ \hline
\multirow{2}{*}{\textbf{\begin{tabular}[c]{@{}c@{}}LLM\\ Backbone\end{tabular}}} & \multicolumn{1}{c|}{\multirow{2}{*}{\textbf{\begin{tabular}[c]{@{}c@{}}Future Instrument\\ (\%)\end{tabular}}}} & \multicolumn{1}{c|}{\multirow{2}{*}{\textbf{\begin{tabular}[c]{@{}c@{}}Future Step\\ (\%)\end{tabular}}}} & \multirow{2}{*}{\textbf{\begin{tabular}[c]{@{}c@{}}Future Phase\\ (\%)\end{tabular}}} & \multicolumn{3}{c}{\textbf{Time (minutes)}}                                                           \\  
                                                                                 & \multicolumn{1}{c|}{}                                                                                           & \multicolumn{1}{c|}{}                                                                                     &                                                                                       & \multicolumn{1}{c|}{\textbf{Phase}} & \multicolumn{1}{c|}{\textbf{Step}} & \textbf{Overall} \\ \hline
\textbf{GPT-2}                                                                   & \multicolumn{1}{c|}{60.29}                                                                                      & \multicolumn{1}{c|}{37.12}                                                                                & 82.5                                                                                  & \multicolumn{1}{c|}{13.04}          & \multicolumn{1}{c|}{8.91}          & 23.13            \\ \hline
\textbf{Qwen3-0.6B}                                                              & \multicolumn{1}{c|}{66.65}                                                                                      & \multicolumn{1}{c|}{33.24}                                                                                & 80.96                                                                                 & \multicolumn{1}{c|}{14.33}          & \multicolumn{1}{c|}{9.38}          & 23.52            \\ \hline
\end{tabular}
}
\end{table*}

\vspace{-0mm}

\begin{table}[!htb]
\centering
\captionsetup{aboveskip=3pt, belowskip=2pt}
\caption{Evaluation of SurgAnt-ViVQA against state of the art image based surgical VQA techniques on the publicly available EndoVis18-VQA benchmark.}
\label{tab:endo_res}
\begin{tabular}{c|c|c|c|c}
\hline
\textbf{Model} & \textbf{BLEU-3} & \textbf{BLEU-4} & \textbf{Rouge-L} & \textbf{Meteor} \\ \hline
VisualBERT              & 70.16 & 67.21 & 80.17 & 75.95 \\
VisualBERT-RM           & 70.33 & 66.40 & 77.58 & 76.06 \\
PitVQA++                & 80.51 & 79.15 & 89.64 & 86.25 \\
SurgAnt-ViVQA (GPT-2)   & 85.82 & 81.29 & 91.66 & 91.15 \\ \hline
\end{tabular}
\end{table}

\vspace{-2mm}

\begin{table}[!htb]
\centering
\captionsetup{aboveskip=3pt, belowskip=2pt}
\caption{Investigates how the number of input video frames (8, 16, 32) affects textual and numeric video QA performance.}
\label{tab:frames}
\begin{tabular}{c|c|c|c|c|c}
\hline
\textbf{No.\ Frames} & \textbf{BLEU-3} & \textbf{BLEU-4} & \textbf{Rouge-L} & \textbf{Meteor} & \textbf{MAE} \\ \hline
8  & 77.02 & 72.38 & 84.94 & 87.05 & 19.1090 \\
16 & 76.00 & 71.23 & 84.39 & 87.01 & 20.5253 \\
32 & 75.97 & 71.15 & 84.61 & 87.15 & 16.1721 \\ \hline
\end{tabular}
\end{table}




\subsection{Ablation Study}

Table \ref{tab:pit_res} compares five cross-attention (CA) variants on the PitVQA-Anticipation benchmark, highlighting the importance of temporal recurrence and gated information flow for multimodal anticipation. Early fusion baselines such as MultistageCA and FusionTok CA outperform LoRA-Enhanced and Dynamic-Gate CA by wide margins (e.g., +30.5 and +34.5 BLEU-4), yet remain constrained by fixed token level fusion. The proposed GRU-Gated Temporal CA achieves state of the art results 88.50/72.38 BLEU-1 to 4, 84.94 ROUGE-L, and 87.05 METEOR representing 71.8 \% and 20.9 \% relative gains in BLEU-4 and METEOR over FusionTok. A lighter 0.6B-parameter version retains 99 \% of BLEU-4 performance, underscoring that improvements stem from temporal gating rather than model scale. Overall, temporal recurrence and gated cross modal fusion emerge as the primary drivers of anticipatory understanding.

Table \ref{tab:endo_res} extends the evaluation to EndoVis18-VQA, where SurgAnt-ViVQA surpasses all prior models across every metric. Relative to VisualBert, it delivers absolute gains of +14.1 BLEU-4, +11.5 ROUGE-L, and +15.2 METEOR (21 \%, 14 \%, and 20 \% improvements), and still exceeds PitVQA++ by +2.1 BLEU-4 and +4.9 METEOR. These results confirm that its surgical scene visual encoder and task-aware cross-modal fusion yield more accurate, contextually grounded answers.

Finally, Table \ref{tab:frames} analyzes the effect of frame sampling on performance. Using 8 frames achieves the best linguistic fluency (BLEU-4 = 72.38, ROUGE-L = 84.94), whereas extending to 32 frames slightly reduces BLEU but lowers numeric error (MAE = 16.17 vs 19.11), indicating improved quantitative reasoning. Thus, concise temporal windows suffice for coherent text generation, suggesting a trade off between fluency and temporal precision depending on task requirements.


\subsection{Qualitative}
\begin{figure}[h]
    \centering
    \includegraphics[width=1\linewidth]{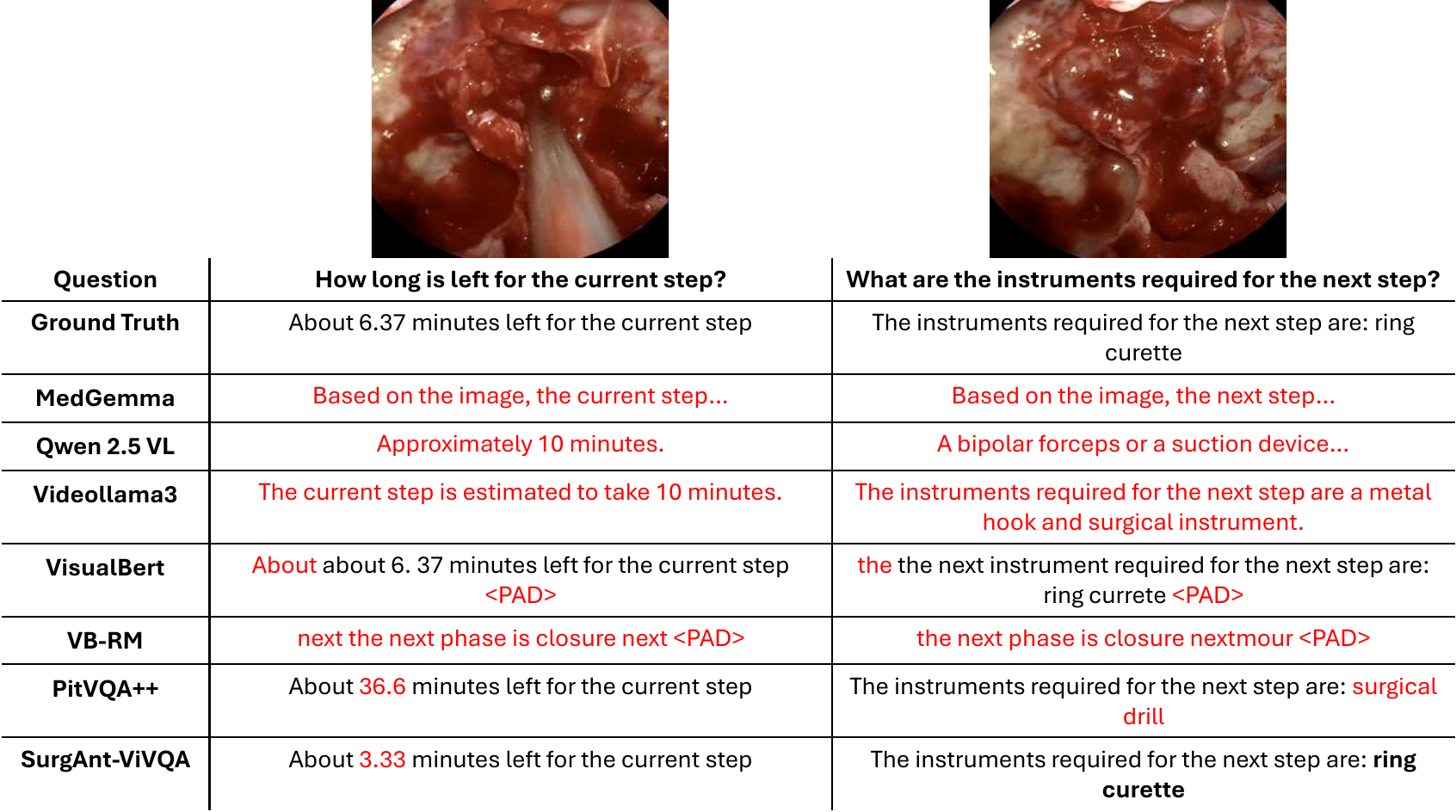}
    \caption{2 examples of sample questions and predictions from PitVQA-Anticipation dataset upon different models. Red text denotes wrong predictions and \texttt{<PAD>} represents the degenerate answers generated.}
    \label{fig:example}
\end{figure}
Fig~\ref{fig:example}. juxtaposes the responses of seven vision language models on a single frame sequence from the PitVQA‑Anticipation benchmark.  Three forward looking questions are posed: \emph{(i)} How long is left for the current step?, and \emph{(ii)} What instruments are required for the next step?. \emph{Generic LMMs.}  MedGemma, Qwen 2.5 VL, LLaMA‑Vision 3.2 and Video-LLaMA3 generate linguistically fluent answers but reveal a lack of domain grounding: fail to provide a concrete time estimate, and list overly general instruments (e.g. bipolar forceps).  These outputs illustrate that large‑scale pre‑training alone is insufficient for step level surgical anticipation. VB and VBRM, which rely on single frame token fusion, collapse to degenerate answers containing \texttt{<PAD>} tokens and therefore add little clinical value.  PitVQA++, a stronger video baseline, correctly detects the current phase but prematurely predicts `operation ended’’ highlighting the difficulty of modelling residual temporal context without explicit gating.\emph{Proposed model.}  In contrast, our \textbf{SurgAnt-ViVQA} equipped with the GRU‑Gated Temporal CA precisely anticipates, predicts the remaining duration within a 3 minute margin, and identifies the \emph{ring curette} as the sole necessary instrument.  The qualitative gap confirms the quantitative gains reported, explicit temporal recurrence coupled with step‑wise gating enables fine‑grained reasoning about forthcoming actions, time horizons, and tool usage capabilities that generic LMMs and token‑level methods lack.  These observations substantiate our claim that surgical anticipation demands task‑specific temporal modelling rather than mere scale or parameter count.

\section{Conclusion}


Our findings demonstrate that transitioning from frame-based VLMs to a temporally attentive video architecture is crucial for advancing surgical VQA from retrospective scene description to proactive decision support. By coupling a bidirectional GRU with adaptive gated cross-attention, \textbf{SurgAnt-ViVQA} outperforms strong image based baselines on the new PitVQA-Anticipation benchmark and generalises effectively to the unrelated EndoVis18-VQA dataset, indicating robust cross procedural transfer. Ablation studies show that temporal recurrence drives most of the performance gain, while adaptive gating enhances precision with minimal parameter cost. Qualitative results confirm that generic LMMs blur future phases and instruments, whereas our model captures both timeline and tooling with clinically relevant granularity. Despite these strengths, PitVQA-Anticipation currently focuses on a single procedure with limited question diversity, and temporal evaluation remains challenging. Future work will extend this benchmark across multiple surgeries with broader question types, explore alternative video embedding strategies, and further assess VLMs’ understanding of temporal dynamics in surgical workflows.

\bmhead{Acknowledgements}
This work was supported by the EPSRC under grant [EP/W00805X/1]. HJM was supported by the NIHR UCLH/UCL Biomedical Research Centre.  
DZK was supported by Cleveland Clinic London for this work.  
Thanks to Medtronic for access to the \textit{Touch Surgery\textsuperscript{TM} Ecosystem} for video recording, annotation, and storage.  
HJM is employed by and holds shares in Panda Surgical Ltd.  
The remaining authors declare no competing interests.

\bmhead{Code Availability}
The source code of this work and PitVQA-Anticipation dataset will be made available at https://github.com/ShreyasDhake/PitVQA-Anticipation.  




\bibliography{sn-bibliography}

\end{document}